%% file: acl_latex.tex
\pdfoutput=1
\documentclass[11pt]{article}

\usepackage[final]{acl}

\usepackage{times}
\usepackage{latexsym}

\usepackage{booktabs}
\usepackage{tabularx}
\usepackage[T1]{fontenc}

\usepackage[utf8]{inputenc}

\usepackage{microtype}

\usepackage{inconsolata}

\usepackage{graphicx}
\usepackage{url}

\newcommand{\ours}{\textsc{SkillAdaptor}}
\usepackage[skins]{tcolorbox}
\usepackage{xcolor}
\usepackage{amsmath}
\usepackage{multirow}
\usepackage{bbm}
\usepackage{booktabs}
\usepackage{colortbl}
\tcbuselibrary{breakable}
\usepackage{amsfonts}
\usepackage{fontawesome5}
\usepackage{enumitem}
\usepackage{comment}
\usepackage[table]{xcolor}
\usepackage{makecell}
\definecolor{gaincolor}{HTML}{28a745}
\definecolor{losscolor}{HTML}{dc3545}

\definecolor{mygray}{gray}{0.9}
\definecolor{myblue}{HTML}{F0FFFF}
\definecolor{refGreenBG}{HTML}{E6F5E6}

\usepackage{calc}
\newlength\myheight
\newlength\mydepth
\settototalheight\myheight{Xygp}
\title{SkillAdaptor: Self-Adapting Skills for LLM Agents from Trajectories}

\author{
    Zhuoyun Yu\textsuperscript{$\spadesuit,\diamondsuit$}\thanks{Equal contribution.},~ 
    Xin Xie\textsuperscript{$\clubsuit$}\footnotemark[1],~ 
    Wuguannan Yao\textsuperscript{$\clubsuit$},~ 
    Chenxi Wang\textsuperscript{$\spadesuit$}, \\
    \textbf{Lei Liang\textsuperscript{$\clubsuit,\diamondsuit$},}~ 
    \textbf{Xiang Qi\textsuperscript{$\clubsuit$},}~ 
    \textbf{Shumin Deng\textsuperscript{$\spadesuit$}\thanks{Corresponding author.}} \\
    \textsuperscript{$\spadesuit$}Zhejiang University, \quad \textsuperscript{$\clubsuit$}Ant Digital Technologies, Ant Group, \\
    \textsuperscript{$\diamondsuit$}Zhejiang University - Ant Group Joint Laboratory of Knowledge Graph \\
    \texttt{\{3220104147, 231sm\}@zju.edu.cn} 
}

\begin{document}
\maketitle

\begin{abstract}
Large language model (LLM) agents increasingly rely on reusable external skills to solve long-horizon interactive tasks. Existing training-free skill adaptation pipelines usually update skills from full trajectories or session-level feedback, which makes failure attribution coarse and often produces unstable or overly broad revisions. We propose \ours{}, a training-free step-level skill adaptation framework with explicit failure attribution, and it can plug into OpenClaw-class agent harnesses. Given a failed trajectory, \ours{} identifies a first actionable fault step, links responsibility to candidate skills, and applies targeted updates under explicit acceptance checks while keeping the backbone frozen. We evaluate on WebShop, PinchBench, and Claw-Eval with Kimi-K2.5, GLM-5, and GPT-5.2. \ours{} improves over no-skill and skill-adaptation baselines on all three suites, with the largest single-metric improvements of $+1.5$ points on PinchBench Avg Score\%, $+1.8$ on Claw-Eval Avg Score, and $+1.7$ on WebShop success rate. These results indicate that step-level attribution supports more stable and auditable training-free skill maintenance\footnote{The code will be released at \url{https://github.com/zjunlp/SkillAdaptor}.}.
\end{abstract}

\section{Introduction}

\begin{figure*}[!t]
  \centering
  \includegraphics[width=\textwidth]{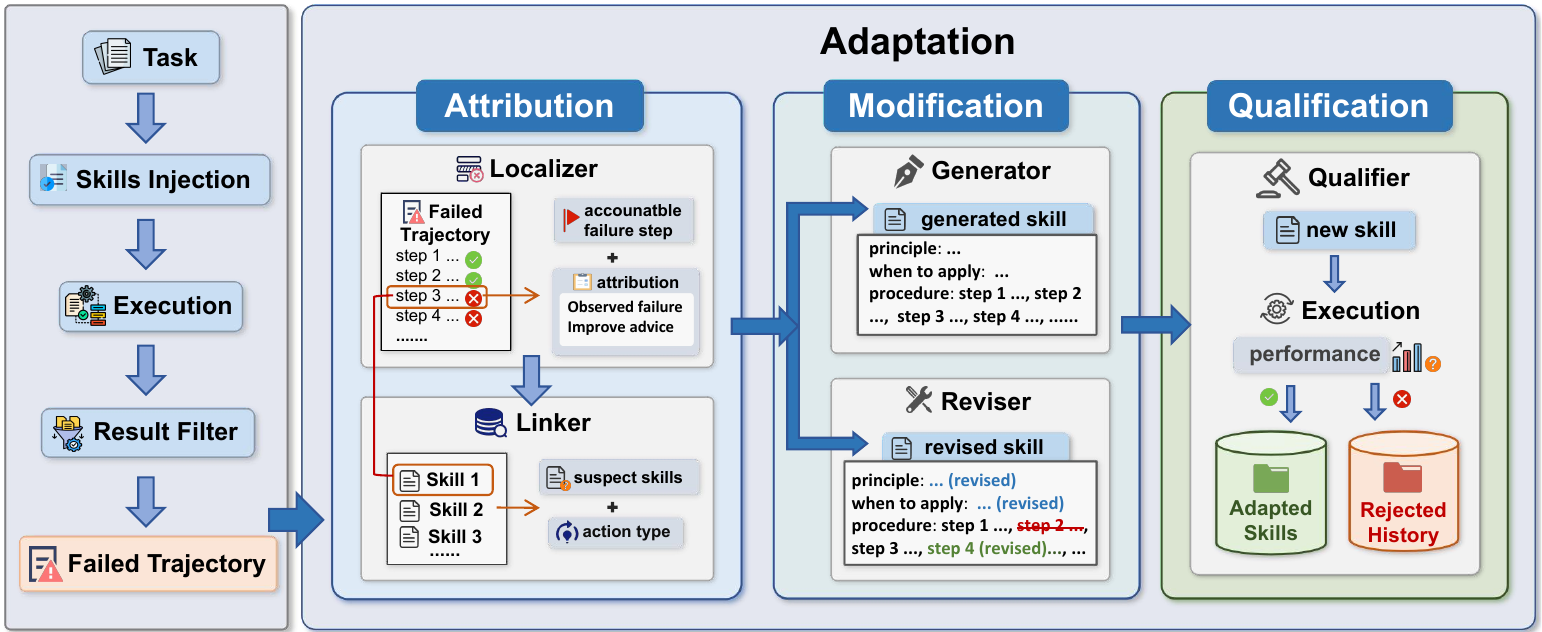}
  \vspace{-7mm}
  \caption{Overview of \ours{} framework.}
  \vspace{-7mm}
  \label{fig:skilladaptor}
\end{figure*}

Large language model (LLM) agents are widely used for long-horizon tasks involving tool use, coding, and web interaction~\cite{yao2023react,schick2023toolformer,patil2023gorilla,wang2023voyager}. To enable continual improvement without retraining the base model, many systems augment frozen LLMs with reusable external skills, such as triggers, action patterns, and verification procedures~\cite{lin2026ahe,openspace_repo,wang2026skillx}.

Recent training-free adaptation methods improve such LLM agents from execution traces through memory distillation~\cite{xu2025amem}, workflow refinement~\cite{wang2024awm}, retrospective feedback~\cite{zhao2024expel}, and iterative skill evolution~\cite{li2026skillforge,alzubi2026evoskill,coevoskills2026}. While effective, most of them adapt skills from trajectory-level outcomes or session-level summaries, using the completed trajectory as the main unit of reflection and revision.

Furthermore, this outcome-level adaptation becomes fragile in long-horizon environments. A failed trajectory often contains many correct intermediate decisions, while the actual failure may result from only a few previous mistakes. When updates are driven mainly by the final outcome, failure signals can be diffused across unrelated steps and skills, leading to overly broad or misdirected revisions. This credit-assignment problem is especially severe when a single early error invalidates many subsequent actions.

To address the aforementioned limitation, we propose \ours{}, a training-free modular skill adaptor that shifts adaptation from trajectory-level reflection to step-level attribution. Given an execution trace, \ours{} localizes the first actionable fault, diagnoses the relevant skill deficiency, and then refines only the relevant skill deficient while leaving the irrelevant skills unchanged. As a modular post-execution adaptation approach, \ours{} can be integrated as a plug-in component for OpenClaw-class agent harnesses~\cite{openclaw_github}.

We evaluate \ours{} on WebShop~\cite{yao2022webshop}, PinchBench~\cite{pinchbench2025}, and Claw-Eval~\cite{claweval2025}. \ours{} consistently improves over training-free adaptation baselines, suggesting that stable agent improvement depends less on expanding skill libraries or reflecting over entire trajectories, and more on accurately attributing failures to the specific steps and skills that caused them.

\section{Preliminaries}
We consider an interactive setting in which an LLM agent solves tasks by acting in an environment. Given a task $q$ sampled from a task distribution $\mathcal{Q}$, the agent produces an execution trajectory
\begin{equation}
    \tau=\{(a_t,o_t)\}_{t=1}^{T},
\end{equation}
where $a_t$ is the action taken by the agent at step $t$ and $o_t$ is the observation. The trajectory is the only evidence available for subsequent fault analysis and skill updates.

To improve task completion without updating model parameters, we maintain a skill collection
\begin{equation}
    K=\{s_1,s_2,\ldots,s_n\},
\end{equation}
where each $s_i$ is a textual skill record carrying a title, a principle, and applicability conditions. The base model is frozen throughout, and only $K$ is revised or appended across iterations.

Our objective is to design the update procedure so that the expected success rate over $\mathcal{Q}$ is improved through $K$ alone
\begin{equation}
    \max_{K}\ \mathbb{E}_{q\sim\mathcal{Q}}
    \left[\operatorname{success}\!\left(\operatorname{Run}(q,K)\right)\right],
\end{equation}
where $\operatorname{Run}(q,K)$ executes the agent on $q$ conditioned on a retrieved subset of $K$, and $\operatorname{success}(\cdot)$ is the benchmark-defined success indicator. This formulation keeps the method training-free and attributes any improvement to updates on $K$.

\section{Method}

\input{tables/exp_main}

\ours{} is a trajectory-driven skill adaptation framework built on a dynamic skill collection $K$. During task execution, the agent retrieves a small set of relevant skills from $K$ and conditions generation on the retrieved context. Failed trajectories are subsequently used to improve  skill collection through Attribution, Modification, and Qualification.

\textbf{Skill Initialization.}
The framework is initialized without any external skill collection. The backbone LLM first executes the task set $\mathcal{Q}$ used for skill adaptation, without any skill or experience augmentation. Successful trajectories are archived as reference traces, from which initial skills are distilled to construct the initial collection $K_0$.

After initialization, all subsequent executions are performed with skill retrieval over $K$. Then failed trajectories from these executions are forwarded to the adaptation stage for skill updating.

\textbf{Skill Injection.}
For each task $q$ with description $d_q$, each skill contains structured fields including title, applicability condition, and behavioral principle. We encode tasks and skills using Qwen3-Embedding-8B denoted by $\phi(\cdot)$. Candidate skills are first retrieved by cosine similarity, truncated to top-$k_{\text{ret}}$ candidates, and finally reranked by the backbone LLM
\begin{equation}
S_q=
\operatorname{Rerank}
\!\left(
\operatorname*{TopK}_{\substack{s\in K}}
\operatorname{sim}\!\big(\phi(d_q),\phi(s)\big),
\, d_q
\right).
\end{equation}

Conditioned on $S_q$, the agent interacts with the
environment and produces a trajectory tuple $(q,\tau,r)$,
where $\tau$ denotes the execution trajectory and
$r$ indicates task success or failure.

\textbf{Skill Adaptation.}
 The Adaptation phrase includes the following three parts. Starting from $K^{(0)}$, we iterate this process for up to $10$ rounds, or until the skill collection remains unchanged for $3$ consecutive rounds. The final collection $K^{\ast}$ is then frozen with no further updates.

\textbf{(1) Attribution.}
Given a failed trajectory, the Localizer predicts the accountable failure step:
\begin{equation}
(t^\ast,\pi)
=\operatorname{Localize}(q,\tau,S_q).
\end{equation}

Here, $t^\ast$ denotes the earliest accountable fault step whose correction is expected to improve the final outcome. $\pi$ is a description of observed failure behavior and improvement suggestion.

The Linker then estimates which retrieved skills are most responsible for the failure:
\begin{equation}
\{(s_j,w_j)\}
=
\operatorname{Link}(q,\tau,t^\ast,S_q).
\end{equation}

The output is a weighted suspect set over retrieved skills, where a larger $w_j$ indicates stronger estimated responsibility for the failure. The Linker additionally predicts an adaptation action $\hat{a}\in\{\textsc{revise},\textsc{generate}\}$. If the failure is attributed to an inappropriate or misleading retrieved skill, the trajectory is routed to skill revision. Otherwise, the failure is treated as an uncovered capability gap due to missing skill coverage and routed to new skill generation.

\textbf{(2) Modification.}
Given the predicted action $\hat{a}$, the Modifier updates the skill collection:
\begin{equation}
K^{+}
=
\operatorname{Modify}(K,t^\ast,\hat{a}).
\end{equation}

where $\operatorname{Modify}$ rewrites the highest-weighted skill and replaces it in $K$ when $\hat{a}=\textsc{revise}$, and synthesizes a new skill from the localized context at $t^\ast$ and appends it when $\hat{a}=\textsc{generate}$.

To control redundancy growth, generated skills are filtered using the same encoder method as retrieval, together with a duplicate threshold $\theta_{\text{dup}}$. Candidate skills whose semantic similarity exceeds the threshold are discarded before insertion.

\textbf{(3) Qualification.}
All candidate updates are qualified respectively before being committed to the skill collection. Given the current collection $K$ and candidate collection $K^{+}$, the framework re-executes tasks under both settings and compares their outcomes:
\begin{equation}
    \Delta=
    \mathbb{E}_{q\sim\mathcal{Q}}
    \bigl[\mathcal{M}(q;K^{+})\bigr]
    -
    \mathbb{E}_{q\sim\mathcal{Q}}
    \bigl[\mathcal{M}(q;K)\bigr]
\end{equation}
where $\mathcal{M}(\cdot)$ denotes the execution feedback metric. The candidate update is accepted only when $\Delta \ge 0$. Otherwise, the modification is discarded and the original collection $K$ is retained unchanged. This qualification stage suppresses harmful updates and stabilizes continual skill adaptation over long interaction horizons.

\section{Experiments}

\input{tables/exp_ablation}

\paragraph{Backbone models.}
We evaluate all methods using three backbone LLMs (Kimi-K2.5, GLM-5, GPT-5.2), and detailed configurations of each backbone model are provided in Appendix~\ref{app:implement}.

\paragraph{Benchmarks and metrics.}
We evaluate on PinchBench~\cite{pinchbench2025}, Claw-Eval~\cite{claweval2025}, and WebShop~\cite{yao2022webshop}. Detailed benchmark descriptions and settings are provided in Appendix~\ref{app:benchmarks}.

\paragraph{Models and baselines.}
We compare with six representative baselines. Details for all baselines are summarized in Appendix~\ref{app:baselines}. All methods use the same backbone models and benchmark settings. Each setting is repeated with three independent runs, and tables report the mean and spread.

\paragraph{Main results.}

As shown in Tables~\ref{tab:main_pinch_claw} and~\ref{tab:main_webshop}, \ours{} consistently improves over reported baselines across all three benchmarks with matched backbones. The clearest improvements appear on WebShop, where gains remain positive across all evaluated models.
On PinchBench and Claw-Eval, improvements are smaller but still positive on all metrics, indicating that the method remains effective in the most open-ended setting.
Overall, finer failure attribution improves the stability of training-free skill adaptation under frozen backbones.
Against the strongest matched skill baseline in each column, the largest gains are $+2.3$ on WebShop Score (GLM-5), $+1.7$ percentage points on WebShop Succ\% (Kimi-K2.5), $+1.5$ on PinchBench Avg Score\% (GLM-5), $+1.8$ on Claw-Eval Avg Score (Kimi-K2.5).

\paragraph{Component ablation.}
\label{sec:ablation}

Table~\ref{tab:ablation} reports Kimi-K2.5 ablation results. Without $\operatorname{Localizer}$ and $\operatorname{Linker}$, WebShop success drops to 28.6\% and score to 35.8, while Claw-Eval Avg Score falls to 74.5, so step-level attribution is needed to target effective revisions.
Without $\operatorname{Qualifier}$, spread rises, including PinchBench $\pm 8.1$ versus $\pm 5.2$ and Claw-Eval Avg Score $\pm 2.7$, and WebShop success falls to 26.3\% with $\pm 2.6$, so removing qualification allows unstable skills to enter retrieval.
When disabling all adaptor modules and keeping only the initial LLM-extracted skill bank, spread stays high on PinchBench ($\pm 7.3$), WebShop success is only $25.3\pm2.5$\%, and WebShop Score is 32.5 versus the 32.1 base model, so gains nearly vanish relative to the full settings. Skills distilled solely from successful trajectories do not reliably transfer across tasks, indicating that effective adaptation depends on precise failure attribution and qualification rather than skill accumulation alone.

\paragraph{Case study.}

\begin{figure*}[t]
  \centering
  \includegraphics[width=\linewidth]{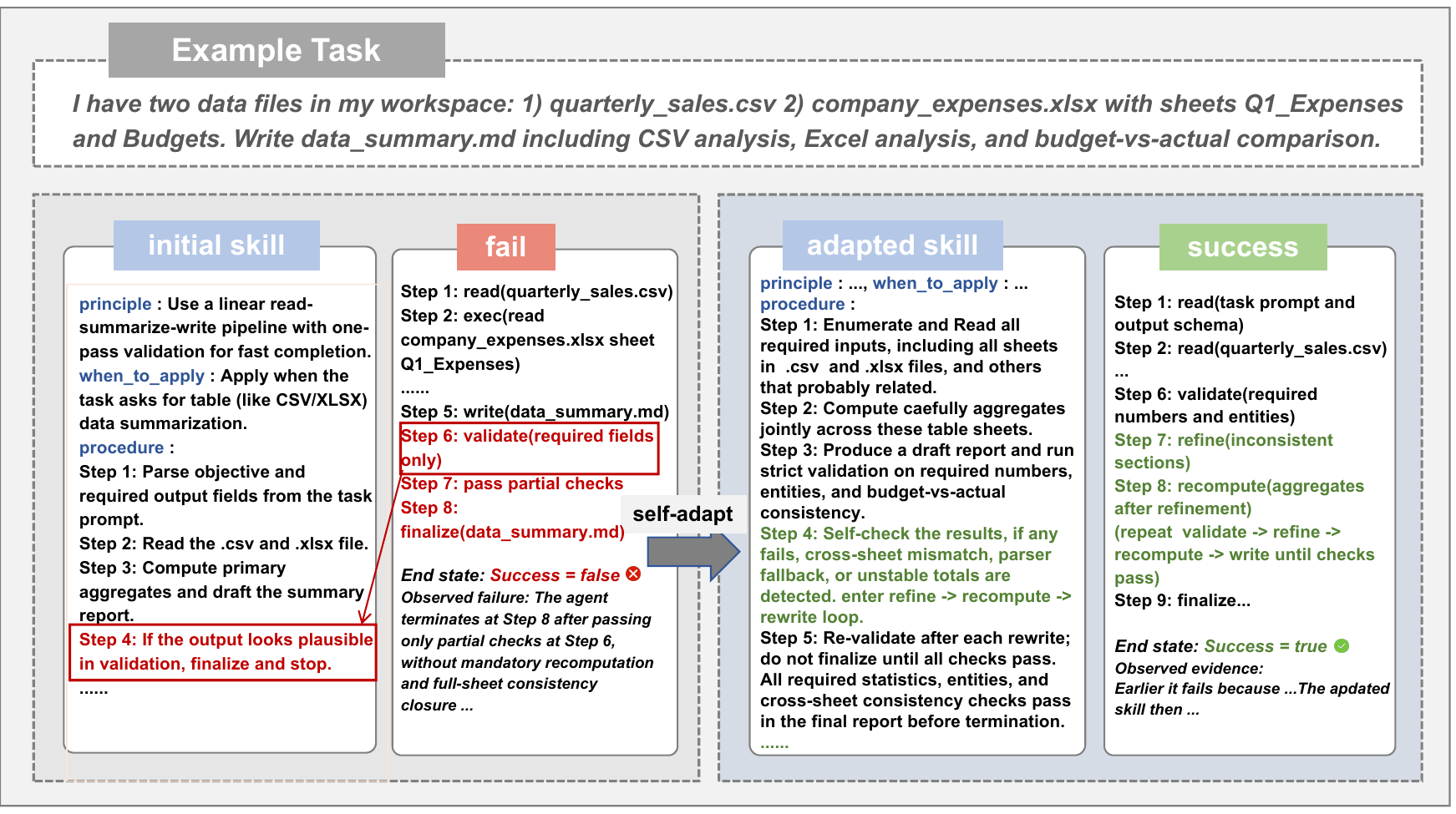}
  \caption{PinchBench task CSV and Excel Data Summarization before (left) and after (right) skill adaptation. The localized failure at interaction step six is mapped to the fourth line of the pre-revision skill, which becomes the target span for revision. The revised card reduces the mismatch between shallow qualification checks and final permission by requiring recomputation before finalization.}
  \label{fig:case-skilladaptor}
\end{figure*}

Trajectory-level failures are often difficult to translate into precise skill updates, because only a small portion of the interaction is directly responsible for the final outcome. We analyze a task on PinchBench, i.e., \texttt{CSV and Excel Data Summarization} (Figure~\ref{fig:case-skilladaptor}), where \ours{} localizes the failure to interaction step six and links it to the fourth procedure of the most relevant skill injected. This attribution converts an otherwise broad trajectory failure into a concrete and directly editable skill span grounded in execution feedback.

The example also reflects a broader pattern across benchmarks. \ours{} is most effective in tasks where failures can be localized to explicit intermediate decisions, such as incorrect parsing, invalid tool arguments, or missing execution steps. These conditions occur most frequently in Data and Code tasks, where execution traces expose a clear procedural structure and a single faulty skill span can often be revised and reused across related problems. More moderate gains are observed in Productivity tasks. In these settings, localized revisions can improve execution ordering, schema usage, or argument selection, but final performance may still be constrained by environment-side rules or external verification requirements. By contrast, improvements are smaller in Research, Memory, and Security tasks, where failures often depend on persistent state, external knowledge, or cross-session context that cannot be fully resolved through localized skill refinement alone.

Overall, \ours{} is most effective when the failure signals are localizable, and execution traces that expose intermediate structure, while less effective when task success depends on external systems or persistent state beyond the scope of a single skill update.

\paragraph{Skill adaptations across adaptor rounds.} \label{sec:roundwise-dynamics} 

\begin{figure}[t] \centering \includegraphics[width=\linewidth]{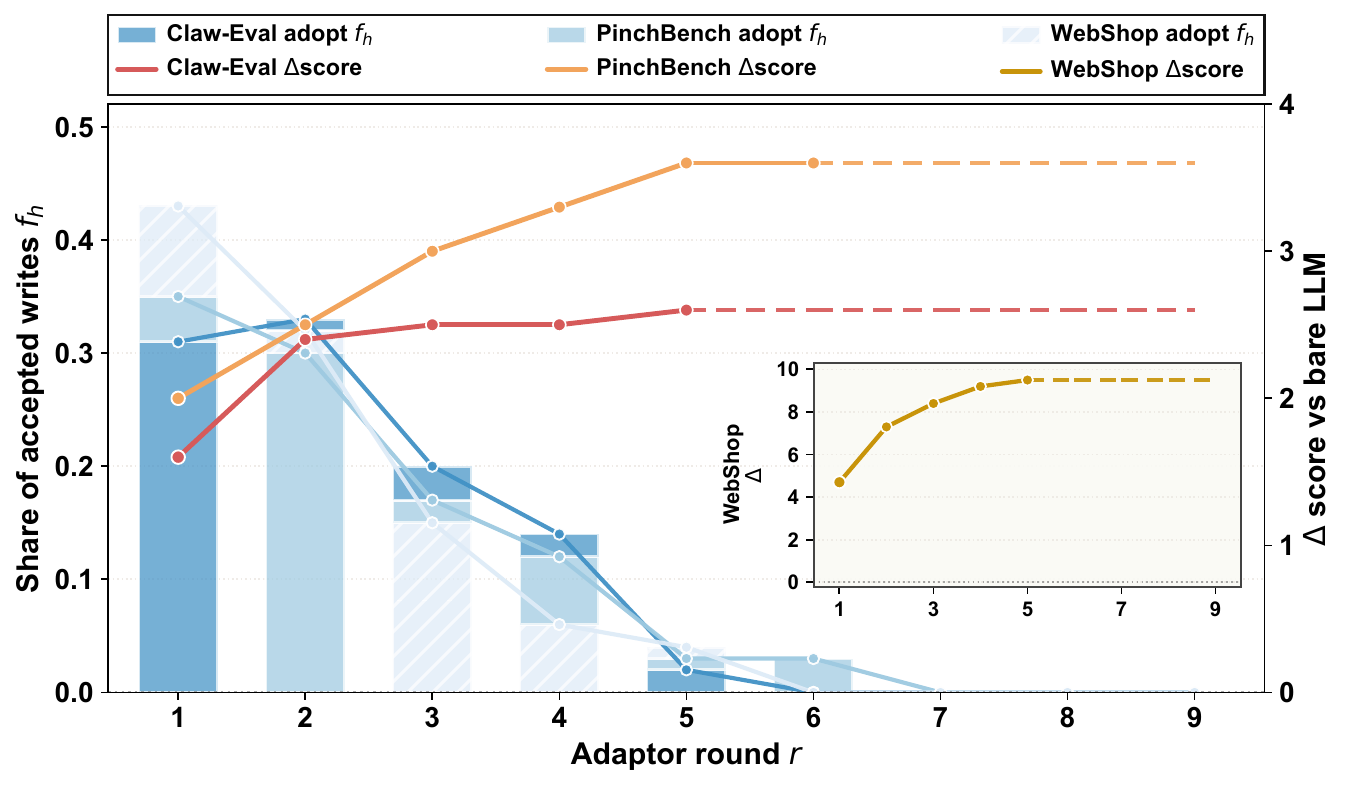}
\caption{Accepted writes and headline-score gains across adaptor rounds on Kimi-K2.5. Left axis is the incremental share $f_h$ of qualifier-approved writes in round~$h$. Right axis shows $\Delta$ in headline score \emph{vs}\ the bare-LLM for Claw-Eval and PinchBench; the inset shows the corresponding WebShop $\Delta$.} 
\label{fig:roundwise-adoption-scores} 
\end{figure}

Figure~\ref{fig:roundwise-adoption-scores} shows the distribution of qualifier-approved skill adaptations across adaptor rounds and the corresponding performance trajectory on three benchmarks. Despite allowing up to ten rounds, the majority of accepted updates concentrate on the first two rounds. WebShop receives roughly 75\% of all accepted writes in rounds 1--2, while Claw-Eval and PinchBench show 64\% and 65\%, respectively. Accepted writes fall to zero after round 5 for WebShop and Claw-Eval, and after round 6 for PinchBench. This early concentration suggests that the dominant recurring failure modes are corrected during the initial adaptation phase, whereas later rounds contribute only incremental refinements. The performance trajectory follows a similar trend, with most gains emerging alongside the earliest accepted updates. The diminishing improvement in later rounds indicates that once major procedural deficiencies are resolved, remaining limitations are increasingly governed by the intrinsic capability of the backbone model rather than missing skill coverage.

\section{Conclusion}
This work presents \ours{}, a training-free adaptor framework that updates external agent skills. Across WebShop, PinchBench, and Claw-Eval, \ours{} consistently improves over no-skill and prior skill-adaptation baselines.
These results suggest that step-level failure attribution enables more precise and efficient agent skill adaptation than trajectory-level updates alone, while remaining lightweight and compatible with existing agent frameworks.

\section*{Limitations}
This study has two limitations.
First, the method is most effective when failures expose observable intermediate signals and required tool dependencies are available, and its performance may weaken under sparse, delayed feedback or missing external interfaces.
Second, while our current experiments cover three public benchmarks, additional evaluation under longer-term deployment settings and broader distribution shifts remains an important direction for future work.

\bibliography{custom}

\clearpage
\appendix

\input{appendix/related_work}
\input{appendix/experimental_setup}
\input{appendix/token_step_stats}


\clearpage
\input{appendix/prompts}

\end{document}

%% file: tables/exp_main.tex
\definecolor{sevRowBg}{HTML}{EEF2FA}
\providecommand{\unc}[2]{#1$^{\text{\scriptsize\color{blue!45!cyan}$\pm$\normalcolor\ensuremath{#2}}}$}

\begin{table*}[t]
    \centering
    \small
    \setlength{\tabcolsep}{4.2pt}
    \resizebox{\textwidth}{!}{
    \begin{tabular}{l c c c @{\hspace{8pt}} cc cc cc}
    \toprule
     & \multicolumn{3}{c}{\textbf{PinchBench}} & \multicolumn{6}{c}{\textbf{Claw-Eval}} \\
    \cmidrule(lr){2-4} \cmidrule(lr){5-10}
     & \textbf{Kimi-K2.5} & \textbf{GLM-5} & \textbf{GPT-5.2} & \multicolumn{2}{c}{\textbf{Kimi-K2.5}} & \multicolumn{2}{c}{\textbf{GLM-5}} & \multicolumn{2}{c}{\textbf{GPT-5.2}} \\
    \cmidrule(lr){5-6} \cmidrule(lr){7-8} \cmidrule(lr){9-10}
    \textbf{Method} & \textbf{Avg Score\%} & \textbf{Avg Score\%} & \textbf{Avg Score\%} & \textbf{Avg Score} & \textbf{Pass@3\%} & \textbf{Avg Score} & \textbf{Pass@3\%} & \textbf{Avg Score} & \textbf{Pass@3\%} \\
    \midrule
    Base Model & \unc{63.6}{8.7} & \unc{70.2}{5.4} & \unc{74.8}{6.1} & \unc{73.2}{1.3} & 74.4\% & \unc{72.8}{1.1} & 72.9\% & \unc{76.0}{1.6} & 77.4\% \\
    OpenSpace & \unc{66.0}{5.9} & \unc{72.3}{4.5} & \unc{75.8}{5.6} & \unc{74.0}{2.2} & 75.9\% & \unc{73.6}{2.1} & 74.4\% & \unc{77.1}{1.8} & 77.4\% \\
    \rowcolor{sevRowBg}\textbf{SkillAdaptor} & \unc{\textbf{67.2}}{5.2} & \unc{\textbf{73.8}}{4.6} & \unc{\textbf{76.6}}{5.1} & \unc{\textbf{75.8}}{1.6} & 77.4\% & \unc{\textbf{75.3}}{1.5} & 75.9\% & \unc{\textbf{77.6}}{1.8} & 78.9\% \\
    \bottomrule
    \end{tabular}
    }
    \vspace{-1mm}
    \caption{Main comparison across PinchBench and Claw-Eval. PinchBench reports Avg Score\%; Claw-Eval reports Avg Score and Pass@3\%.}
    \vspace{-1mm}
    \label{tab:main_pinch_claw}
\end{table*}

\begin{table}[t]
    \centering
    \small
    \setlength{\tabcolsep}{4.2pt}
    \resizebox{\columnwidth}{!}{
    \begin{tabular}{l cc cc cc}
    \toprule
    \multicolumn{7}{c}{\textbf{WebShop}} \\
    \midrule
     & \multicolumn{2}{c}{\textbf{Kimi-K2.5}} & \multicolumn{2}{c}{\textbf{GLM-5}} & \multicolumn{2}{c}{\textbf{GPT-5.2}} \\
    \cmidrule(lr){2-3} \cmidrule(lr){4-5} \cmidrule(lr){6-7}
    \textbf{Method} & \textbf{Score} & \textbf{Succ\%} & \textbf{Score} & \textbf{Succ\%} & \textbf{Score} & \textbf{Succ\%} \\
    \midrule
    Base Model & \unc{32.1}{1.4} & \unc{24.6}{1.5} & \unc{33.8}{0.9} & \unc{25.3}{2.0} & \unc{36.2}{1.0} & \unc{25.0}{1.6} \\
    A-Mem & \unc{30.3}{1.6} & \unc{22.6}{2.1} & \unc{30.9}{0.8} & \unc{20.0}{1.8} & \unc{33.7}{1.2} & \unc{21.6}{2.4} \\
    AWM & \unc{34.0}{0.9} & \unc{25.3}{1.4} & \unc{34.5}{1.6} & \unc{25.6}{1.7} & \unc{36.5}{1.3} & \unc{28.3}{1.9} \\
    ExpeL & \unc{35.7}{0.6} & \unc{26.0}{1.1} & \unc{35.2}{1.0} & \unc{27.3}{1.2} & \unc{37.0}{0.7} & \unc{29.6}{1.4} \\
    EvoSkill & \unc{40.4}{0.3} & \unc{31.3}{0.8} & \unc{41.6}{0.6} & \unc{32.6}{0.8} & \unc{43.1}{0.5} & \unc{33.0}{1.1} \\
    \rowcolor{sevRowBg}\textbf{SkillAdaptor} & \unc{\textbf{41.6}}{0.5} & \unc{\textbf{33.0}}{1.1} & \unc{\textbf{43.9}}{0.7} & \unc{\textbf{33.3}}{0.4} & \unc{\textbf{44.8}}{0.4} & \unc{\textbf{34.3}}{1.2} \\
    \bottomrule
    \end{tabular}
    }
    \vspace{-1mm}
    \caption{Results on WebShop.}
    \vspace{-1mm}
    \label{tab:main_webshop}
\end{table}

%% file: tables/exp_ablation.tex
\providecommand{\unc}[2]{#1$^{\text{\scriptsize\color{blue!45!cyan}$\pm$\normalcolor\ensuremath{#2}}}$}

\begin{table*}[h]
    \centering
    \small
    \setlength{\tabcolsep}{3.8pt}
    \resizebox{\textwidth}{!}{
    \begin{tabular}{l cc c cc}
    \toprule
    \textbf{Configuration} &
    \multicolumn{2}{c}{\textbf{WebShop}} &
    \textbf{PinchBench} &
    \multicolumn{2}{c}{\textbf{Claw-Eval}} \\
    \cmidrule(lr){2-3} \cmidrule(lr){4-4} \cmidrule(lr){5-6}
     & \textbf{Score} & \textbf{Succ\%} & \textbf{Avg Score\%} & \textbf{Avg Score} & \textbf{Pass@3\%} \\
    \midrule
    (0) Base model (no skill bank) &
    \unc{32.1}{1.4} & \unc{24.6}{1.5} &
    \unc{63.6}{8.7} &
    \unc{73.2}{1.3} & 74.4\% \\
    \midrule
    (1) \ours{} (full) &
    \unc{\textbf{41.6}}{0.8} & \unc{\textbf{33.0}}{1.0} &
    \unc{\textbf{67.2}}{5.2} &
    \unc{\textbf{75.8}}{1.6} & \textbf{77.4\%} \\
    (2) w/o $\operatorname{Localizer}$ \& $\operatorname{Linker}$ &
    \unc{35.8}{0.9} & \unc{28.6}{1.5} &
    \unc{65.3}{6.8} &
    \unc{74.5}{1.4} & 75.9\% \\
    (3) w/o qualifier gate &
    \unc{34.0}{2.1} & \unc{26.3}{2.6} &
    \unc{65.8}{8.1} &
    \unc{74.2}{2.7} & 75.9\% \\
    (4) w/o $\operatorname{Localizer}$, $\operatorname{Linker}$, $\operatorname{Qualifier}$ (initial skills only) &
    \unc{32.5}{1.0} & \unc{25.3}{2.5} &
    \unc{64.1}{7.3} &
    \unc{73.9}{1.8} & 74.4\% \\
    \bottomrule
    \end{tabular}
    }
    \caption{Component ablations on Kimi-K2.5 (same harness and metrics as Table~\ref{tab:main_pinch_claw}\&\ref{tab:main_webshop}). Row~(2) removes $\operatorname{Localizer}$ and $\operatorname{Linker}$, row~(3) removes  $\operatorname{Qualifier}$, row~(4) uses only round-1 extracted skills without further attributed edits.}
    \label{tab:ablation}
\end{table*}

%% file: appendix/related_work.tex
\section{Related Work}

\textbf{Long-horizon agent systems and capability decomposition}
Large language model agents are increasingly applied to long-horizon interactive tasks such as WebShop~\cite{yao2022webshop}, software terminals, and tool-augmented workflows~\cite{yao2023react, schick2023toolformer, patil2023gorilla, hong2024metagpt}. Recent work studies how reasoning, action, memory, and tool usage can be integrated into unified execution loops~\cite{shinn2023reflexion, park2023generativeagents, wang2023voyager}. Existing systems are commonly organized around interaction, memory, and skills. While interaction and memory have been extensively studied through tool augmentation and retrieval mechanisms, skill evolution in long-horizon environments remains less explored, especially under concrete execution failures.

\textbf{Self-adapting skill methods}
Recent work explores how skills can be extracted and refined from agent experiences. Ctx2Skill~\cite{si2026ctx2skill} studies context-aware retrieval and large-scale skill infrastructure. SkillForge~\cite{li2026skillforge} performs failure analysis and minimal skill revision from execution traces. EvoSkill~\cite{alzubi2026evoskill} studies skill discovery under distribution shift, while CoEvoSkills~\cite{coevoskills2026} introduces iterative generation and verification for skill refinement. 
SkillClaw~\cite{skillclaw2026} extends skill adaptation to deployment settings through cross-session feedback aggregation. SkillsVote~\cite{skillsvote2026} further studies lifecycle governance for reusable agent skills, including collection, recommendation, attribution, and evidence-gated evolution in large-scale skill ecosystems. SkillX~\cite{wang2026skillx} and SkillOS~\cite{ouyang2026skillos} further study hierarchical skill organization and feedback-driven curation.
SkillOpt~\cite{yang2026skillopt} formulates skill editing as a controllable evolution process with validation-gated textual updates to compact skill artifacts.
Library Drift~\cite{zhang2026librarydrift} studies failure modes in self-evolving skill libraries arising from semantic and distributional drift over time.

Most existing approaches operate at the trajectory or session level, where failures are aggregated before updates are applied. As a result, they often lack precise localization of the execution step responsible for downstream errors. Several methods further rely on multi-agent coordination or iterative verification pipelines for adaptation, yet our method adopts a lightweight single-agent framework, enabling direct and efficient skill adaptation.

\textbf{Harness systems and skill adaptation}
Another line of work studies how external orchestration systems stabilize agent execution. AHE~\cite{lin2026ahe} and OpenSpace~\cite{openspace_repo} improve execution reliability through workflow organization and controlled tool interaction. Complementary studies investigate skill retrieval and reuse at scale. Skill-Usage~\cite{liu2026skillusage} and SkillsBench~\cite{skillsbench2026} evaluate skill effectiveness across tasks. Huang et al.~\cite{huang2026rawexperience} conduct a systematic study
of the full trajectory-to-skill lifecycle, introducing extraction efficacy and target evolvability metrics to disentangle skill quality from model adaptability.

These studies improve execution robustness and skill reuse, but they do not directly address how reusable skills should be updated after concrete execution failures. Our work instead focuses on step-level failure attribution and fine-grained skill evolution under long-horizon interaction.

%% file: appendix/experimental_setup.tex
\section{Experimental Details}
\label{app:setup}

\subsection{Implementation Details}
\label{app:implement}

All stages of skill adaptation, including initialization rollout, failure localization, responsibility attribution, skill revision, skill generation, and qualification, are performed using the same backbone LLM configured for the current experiment setting.

\paragraph{Skill Initialization, Injection and Adaptation.}
Skills are represented using the SKILL.md format adopted by OpenClaw-style agents, where each skill contains lightweight structured metadata together with natural-language behavioral instructions and optional execution guidance.

For skill extraction, generation, and revision, we use a relatively high temperature ($0.9$) to encourage diverse reasoning traces and broaden coverage of failure-induced behaviors.

We use Qwen3-Embedding-8B for dense retrieval over the skill collection.
At inference time, candidate skills are first filtered by cosine similarity using a minimum threshold of $0.45$, after which the top-$10$ candidates are reranked by the backbone LLM conditioned on the current task description. 

To suppress redundancy accumulation during continual adaptation, newly generated or revised skills are semantically compared against the existing collection using the same embedding model.
Candidate updates whose cosine similarity exceeds the duplicate threshold $\theta_{\text{dup}}=0.95$ are discarded before insertion.

Skill adaptation proceeds for at most $10$ rounds, with early stopping triggered if the skill collection remains unchanged for $3$ consecutive rounds.

\paragraph{Execution.}
Across all benchmarks, we set the temperature to $0$ for execution. For WebShop, all methods are evaluated in the official WebShop Gym-based environment and the standard evaluator released by the benchmark authors. For PinchBench and Claw-Eval, experiments are conducted using the benchmark-provided OpenClaw runtime and task configurations.
We keep the default sandbox, tool interfaces, execution settings, grading pipeline, and aggregation rules unchanged, and vary only the backbone LLM or adaptation method across experiments.

During skill iteration, the adaptor only has access to execution trajectories and their success or failure signals. It does not observe internal evaluator logic, reward decomposition, or any benchmark-specific scoring details. All sandbox settings, tool interfaces, execution configurations, grading pipelines, and aggregation rules are kept unchanged, while we vary only the backbone LLM or the adaptation method across experiments.

\subsection{Benchmarks}
\label{app:benchmarks}

\paragraph{WebShop}
WebShop\citep{yao2022webshop} is an interactive web navigation benchmark for multi-step online shopping tasks. We follow the data split of \citet{jung2025coevolving}, with 1,624 training instances, and a test set of 200 instructions. Our tables report \textbf{task score} and \textbf{success rate} on the test set, with identical environment settings across methods.

\paragraph{PinchBench}
PinchBench\citep{pinchbench2025} is an OpenClaw\citep{openclaw_github} benchmark for long-horizon tool-mediated agents. The task set covers coding and DevOps scripts, spreadsheets, and log analysis, meeting transcripts, email and calendar workflows, and web research briefs. We report \textbf{Avg Score\%} in the main table.

\paragraph{Claw-Eval}
Claw-Eval\citep{claweval2025} is an OpenClaw\citep{openclaw_github} benchmark for open-ended agent evaluation. Tasks span single-turn office and ops workflows, finance and ticket triage, and multi-turn advisory dialogues with a simulated user. We use the Overall Dimension of the official leaderboard, which follows the benchmark’s standard aggregation protocol and consists of General tasks and Multi-turn tasks. We report \textbf{Avg Score} and \textbf{Pass@3} in the main table.

\subsection{Baselines}
\label{app:baselines}

Unless stated otherwise, each baseline sits on the same training-free base LLM as our method, under the same experiment settings. External skills or memory are applied in whatever module the baseline defines.

\paragraph{Base model}
This method runs the policy with no external skills or experience injected. We evaluate test tasks on Kimi-K2.5, GLM-5, and GPT-5.2. 

\paragraph{A-Mem}
A-Mem\citep{xu2025amem} implements agentic memory.
The model chooses what to store, retrieve, and revise in a persistent store rather than a fixed scratchpad.
We follow the public A-Mem recipe, including the benchmark prompts and harness assumptions described in the original release.

\paragraph{Agent Workflow Memory}
Agent Workflow Memory\citep{wang2024awm}, abbreviated AWM, distills reusable workflows from past trajectories and retrieves them on new tasks with lightweight string matching.
We reuse the wiring prescribed in our benchmark runbook with identical tool calls, step limits, and scoring.

\paragraph{ExpeL}
ExpeL\citep{zhao2024expel} extracts qualitative lessons from high- versus low-reward rollouts and injects them as free-text hints at test time.
Its artifacts read as retrospective critiques rather than explicit workflow templates.

\paragraph{EvoSkill}
EvoSkill\citep{alzubi2026evoskill} performs iterative skill discovery from traces and maintains a skill pool over time.
We run it in the same training-free regime as our other skill-bank baselines and report WebShop numbers in Table~\ref{tab:main_webshop}.

\paragraph{OpenSpace}
OpenSpace\citep{openspace_repo} is a training-free agent system that ships skill libraries and iterative skill refinement loops on top of an OpenClaw-class harness.
The released codebase targets long-horizon desktop-style agents rather than a single fixed policy template.
We submit runs through the same PinchBench and Claw-Eval channels as the base model so that scores stay comparable under the OpenClaw evaluation stack\citep{openclaw_github}.

%% file: appendix/token_step_stats.tex
\section{Input Tokens and Interaction Steps}
\label{app:token-step-stats}

\begin{figure}[t]
  \centering
  \includegraphics[width=\linewidth]{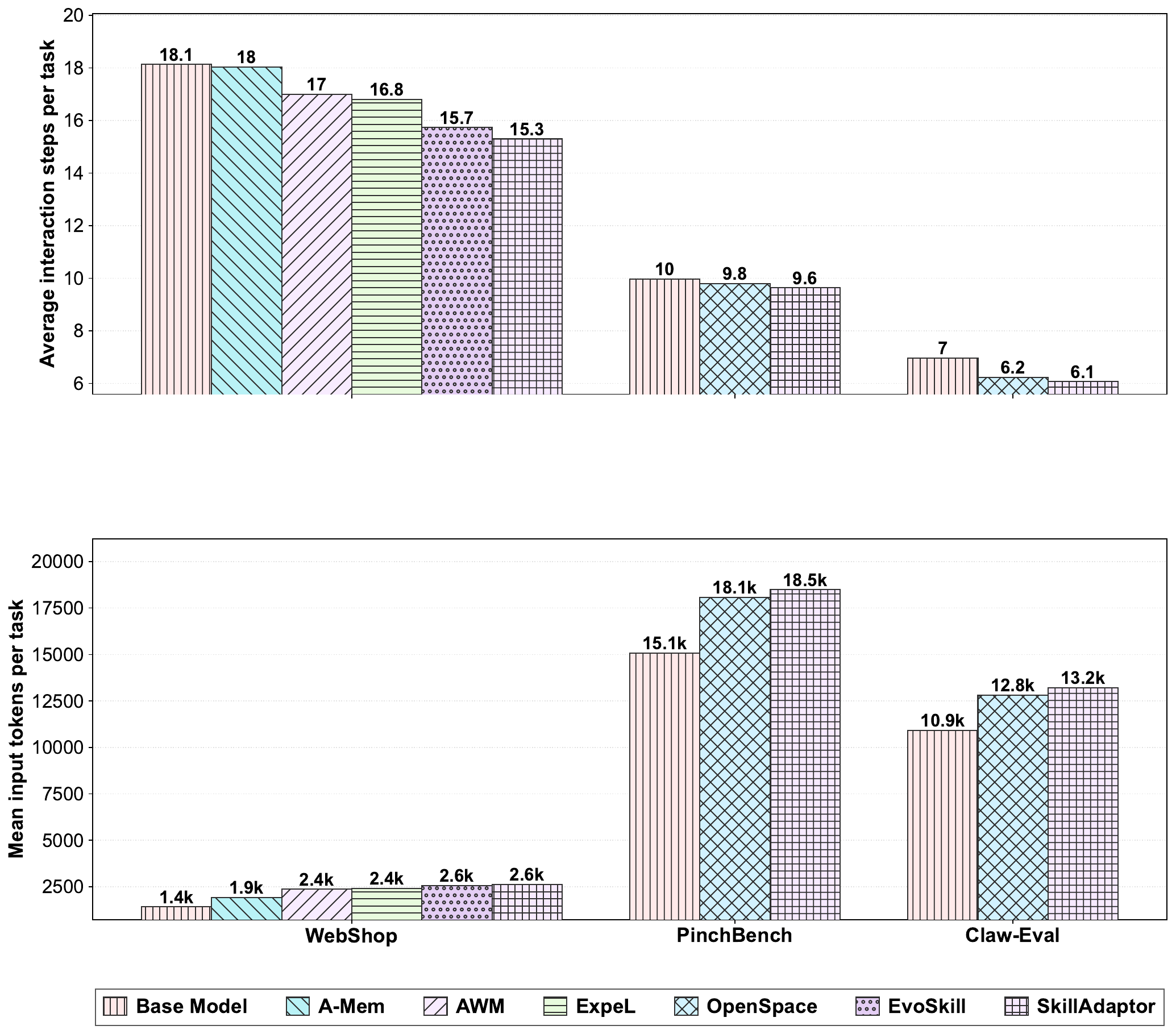}
  \caption{Overview of mean interaction steps (top) and mean input tokens (bottom) per task across the three benchmarks, for the methods listed in Tables~\ref{tab:appendix-token-stats} and~\ref{tab:appendix-steps-stats}. SkillAdaptor increases measured prompt size while reducing mean steps on PinchBench and Claw-Eval relative to Base. WebShop stays in a lower token band with smaller absolute shifts.}
  \label{fig:appendix-steps-tokens-en}
\end{figure}

Retrieving and injecting skills add procedural descriptions to the prompt in each task, which inflates input tokens. This effect is more pronounced on PinchBench and Claw-Eval, where tool usage patterns and multi-step reasoning chains lead to repeated exposure of skill-related context across interactions. For Kimi-K2.5, mean input tokens increase from 13.9k to 16.9k on PinchBench and from 11.5k to 13.9k on Claw-Eval, while WebShop increases from 1.5k to 2.7k, remaining within a lower token range.

Despite the increased per-step context, the average number of interaction steps consistently decreases on the two desktop benchmarks. Mean steps decrease from 10.4 to 9.8 on PinchBench and from 7.3 to 5.8 on Claw-Eval, with a similar reduction from 18.9 to 15.5 on WebShop. This pattern suggests that skill augmentation shifts part of the decision burden from multi-step interaction to in-context reasoning at each step, reducing the need for repeated exploration and correction at the trajectory level.

Importantly, the reduction in interaction steps does not directly imply lower overall computational cost, but instead reflects a redistribution of computation from environmental interaction to richer per-step context conditioning. As a result, token usage and interaction length should be interpreted jointly, as they capture different aspects of the same adaptation process rather than independent efficiency measures.

\input{tables/appen_token_stats}
\input{tables/appen_steps_stats}

%% file: tables/appen_token_stats.tex
\begin{table*}[t]
    \centering
    \small
    \setlength{\tabcolsep}{4.2pt}
    \definecolor{sevRowBg}{HTML}{EEF2FA}
    \begin{tabularx}{\textwidth}{@{}>{\raggedright\arraybackslash}X>{\centering\arraybackslash}X>{\centering\arraybackslash}X>{\centering\arraybackslash}X@{}}
    \toprule
    \multicolumn{4}{>{\hsize=4.0\hsize\relax\centering\arraybackslash}X@{}}{\textbf{WebShop}} \\
    \midrule
    \textbf{Method} & \textbf{Kimi-K2.5} & \textbf{GLM-5} & \textbf{GPT-5.2} \\
    \midrule
    Base Model & 1507 & 1308 & 1440 \\
    A-Mem & 1968 & 1810 & 1985 \\
    AWM & 2364 & 2229 & 2507 \\
    ExpeL & 2316 & 2311 & 2592 \\
    EvoSkill & 2585 & 2450 & 2663 \\
    \rowcolor{sevRowBg}\textbf{SkillAdaptor} & 2651 & 2516 & 2690 \\
    \midrule
    \multicolumn{4}{>{\hsize=4.0\hsize\relax\centering\arraybackslash}X@{}}{\textbf{PinchBench}} \\
    \midrule
    \textbf{Method} & \textbf{Kimi-K2.5} & \textbf{GLM-5} & \textbf{GPT-5.2} \\
    \midrule
    Base Model & 13927 & 15039 & 16284 \\
    OpenSpace & 16207 & 19052 & 18941 \\
    \rowcolor{sevRowBg}\textbf{SkillAdaptor} & 16902 & 19954 & 18607 \\
    \midrule
    \multicolumn{4}{>{\hsize=4.0\hsize\relax\centering\arraybackslash}X@{}}{\textbf{Claw-Eval}} \\
    \midrule
    \textbf{Method} & \textbf{Kimi-K2.5} & \textbf{GLM-5} & \textbf{GPT-5.2} \\
    \midrule
    Base Model & 11470 & 10515 & 10763 \\
    OpenSpace & 13684 & 12042 & 12720 \\
    \rowcolor{sevRowBg}\textbf{SkillAdaptor} & 13933 & 12769 & 12917 \\
    \bottomrule
    \end{tabularx}
    \caption{Input token statistics. Values are mean input tokens per task execution.}
    \label{tab:appendix-token-stats}
\end{table*}

%% file: tables/appen_steps_stats.tex
\begin{table*}[t]
    \centering
    \small
    \setlength{\tabcolsep}{4.2pt}
    \definecolor{sevRowBg}{HTML}{EEF2FA}
    \begin{tabularx}{\textwidth}{@{}>{\raggedright\arraybackslash}X>{\centering\arraybackslash}X>{\centering\arraybackslash}X>{\centering\arraybackslash}X@{}}
    \toprule
    \multicolumn{4}{>{\hsize=4.0\hsize\relax\centering\arraybackslash}X@{}}{\textbf{WebShop}} \\
    \midrule
    \textbf{Method} & \textbf{Kimi-K2.5} & \textbf{GLM-5} & \textbf{GPT-5.2} \\
    \midrule
    Base Model & 18.9 & 18.4 & 17.1 \\
    A-Mem & 17.9 & 19.2 & 17.0 \\
    AWM & 17.5 & 17.1 & 16.4 \\
    ExpeL & 16.6 & 17.9 & 15.9 \\
    EvoSkill & 15.8 & 16.5 & 14.9 \\
    \rowcolor{sevRowBg}\textbf{SkillAdaptor} & 15.5 & 15.3 & 15.1 \\
    \midrule
    \multicolumn{4}{>{\hsize=4.0\hsize\relax\centering\arraybackslash}X@{}}{\textbf{PinchBench}} \\
    \midrule
    \textbf{Method} & \textbf{Kimi-K2.5} & \textbf{GLM-5} & \textbf{GPT-5.2} \\
    \midrule
    Base Model & 10.4 & 10.2 & 9.3 \\
    OpenSpace & 10.1 & 9.6 & 9.7 \\
    \rowcolor{sevRowBg}\textbf{SkillAdaptor} & 9.8 & 9.2 & 9.9 \\
    \midrule
    \multicolumn{4}{>{\hsize=4.0\hsize\relax\centering\arraybackslash}X@{}}{\textbf{Claw-Eval}} \\
    \midrule
    \textbf{Method} & \textbf{Kimi-K2.5} & \textbf{GLM-5} & \textbf{GPT-5.2} \\
    \midrule
    Base Model & 7.3 & 6.9 & 6.7 \\
    OpenSpace & 6.0 & 6.6 & 6.1 \\
    \rowcolor{sevRowBg}\textbf{SkillAdaptor} & 5.8 & 6.4 & 6.0 \\
    \bottomrule
    \end{tabularx}
    \caption{Execution step statistics. Values are mean interaction steps per task execution.}
    \label{tab:appendix-steps-stats}
\end{table*}

%% file: appendix/prompts.tex
\onecolumn
\newpage
\section{Prompt Templates}
\label{app:prompts_details}

This appendix presents the core prompt templates used in the training-free
\ours{} adaptor pipeline. We report compact but complete prompt skeletons for
failure localization, responsibility linking, skill revision, skill generation,
and qualification gating. All templates are written in English and intended for
top-tier conference style reproducibility.


\begin{figure*}[!htbp]
\centering
\begin{tcolorbox}[enhanced, breakable, title=Shared Constraint Template]
\textbf{Role.}
You are an expert agent debugger. Output must be deterministic,
evidence-grounded, and directly actionable.

\textbf{Global constraints.}
\begin{itemize}[leftmargin=*]
  \item Do not fabricate facts; use only trajectory evidence.
  \item If repeated actions show no progress, explicitly flag loop risk.
  \item Never propose runtime package installation or environment creation.
  \item Flag \textbf{same-tool same-parameter} calls repeated $\geq 3$ times
  without meaningful intermediate operations (excluding benign repeats such as
  Read/Grep/Glob/TodoWrite).
  \item Evaluate sensitive-path risk only from executable parameters, not from
  natural-language path mentions.
  \item In shell contexts, treat \texttt{sudo} and \texttt{su -} as
  privilege-escalation command tokens; avoid false positives in non-command
  arguments.
  \item Reject destructive operations targeting root or system-critical
  directories; enforce workspace-scoped operations.
\end{itemize}
\end{tcolorbox}
\caption{Shared safety and quality constraints applied across all stages.}
\label{prompt:shared:constraints}
\end{figure*}


\begin{figure*}[!htbp]
\centering
\begin{tcolorbox}[enhanced, breakable, title=Fault Localizer Prompt]
\textbf{Task.}
Analyze a failed trajectory, extract a candidate fault chain, and return one
primary fault step for this round.

\textbf{Input fields.}
Task description, trajectory summary, recent critical steps, and fault-type
definitions.

\textbf{Instructions.}
\begin{itemize}[leftmargin=*]
  \item Extract 2--4 candidate fault steps that are most likely to explain the
  failure, ranked by evidence strength.
  \item Select one primary step $t^\ast$ from the candidate chain for the
  current revision attempt.
  \item Distinguish \texttt{skill\_wrong} (an existing skill misguided the
  agent) and \texttt{skill\_missing} (no existing skill covered the situation).
  \item If repetitive ineffective actions appear, include loop-prevention
  guidance in the improvement principle.
\end{itemize}

\textbf{Output format (strict).}
\begin{itemize}[leftmargin=*]
  \item \texttt{fault\_chain}: ranked step ids (1-based), e.g., [5,7,9]
  \item \texttt{t\_star}: primary step selected from \texttt{fault\_chain}
  \item \texttt{improvement\_principle}: one concise correction statement
  \item \texttt{fault\_type}: one of predefined classes
  \item \texttt{reason}: brief evidence-grounded explanation
\end{itemize}
\end{tcolorbox}
\caption{Template for fault-chain localization and primary-step selection in failed trajectories.}
\label{prompt:localizer}
\end{figure*}


\begin{figure*}[!htbp]
\centering
\begin{tcolorbox}[enhanced, breakable, title=Skill Linker Prompt]
\textbf{Task.}
Assign responsibility weights from a localized failure to active skill
candidates.

\textbf{Input fields.}
Fault context, wrong action, improvement principle, and candidate skills
(title/description/trigger snippets).

\textbf{Attribution rules.}
\begin{itemize}[leftmargin=*]
  \item Use weighted evidence from direct instruction match, context mismatch,
  omission, and misleading wording.
  \item Typical ranges: 0.8--1.0 (fully responsible), 0.5--0.7 (partially
  responsible), 0.2--0.4 (weakly related), 0.0--0.1 (irrelevant).
  \item If no skill is meaningfully responsible, return an empty attribution
  list.
\end{itemize}

\textbf{Output format (JSON).}
\texttt{\{"attributions":[\{"skill\_id":"...", "weight":0.75,
"reason":"..."\}]\}}
\end{tcolorbox}
\caption{Template for step-level responsibility attribution.}
\label{prompt:linker}
\end{figure*}


\begin{figure*}[!htbp]
\centering
\begin{tcolorbox}[enhanced, breakable, title=Skill Reviser Prompt]
\textbf{Task.}
Perform minimal, targeted edits to an existing skill after a linked failure.

\textbf{Revision policy.}
\begin{itemize}[leftmargin=*]
  \item Preserve working parts and prefer additive edits.
  \item Add preconditions, disambiguation, negative examples, and qualification
  checks only where evidence supports the change.
  \item Include explicit anti-loop and safety guards when failure traces show
  repetitive behavior or risky shell actions.
\end{itemize}

\textbf{Output format (JSON).}
\begin{itemize}[leftmargin=*]
  \item Use the same skill schema as the generator and wrap it with mode metadata.
  \item \texttt{update\_mode = revise\_existing}
  \item \texttt{target\_skill\_id = <existing skill id>}
  \item \texttt{skill\_profile} with strict fields:
  \item \texttt{title}, \texttt{principle}, \texttt{when\_to\_apply}
  \item \texttt{procedure} (3--6 deterministic steps)
  \item \texttt{qualification\_criteria}
  \item \texttt{negative\_example\{what\_not\_to\_do, why\_it\_fails\}}
\end{itemize}
\end{tcolorbox}
\caption{Template for minimal skill revision under explicit evidence.}
\label{prompt:reviser}
\end{figure*}


\begin{figure*}[!htbp]
\centering
\begin{tcolorbox}[enhanced, breakable, title=Skill Generator Prompt]
\textbf{Task.}
Convert failure evidence into a reusable new skill when no existing skill can
be safely revised.

\textbf{Generation policy.}
\begin{itemize}[leftmargin=*]
  \item Extract reusable principles, not one-off task hacks.
  \item Define precise triggers as \emph{observation pattern + state
  condition}.
  \item Provide executable procedures with verification checkpoints.
  \item Include explicit negative examples and stop conditions.
\end{itemize}

\textbf{Output format (JSON).}
\begin{itemize}[leftmargin=*]
  \item Use the same skill schema as the reviser and wrap it with mode metadata.
  \item \texttt{update\_mode = generate\_new}
  \item \texttt{target\_skill\_id = null}
  \item \texttt{skill\_profile} with strict fields:
  \item \texttt{title}, \texttt{principle}, \texttt{when\_to\_apply}
  \item \texttt{procedure} (3--6 deterministic steps)
  \item \texttt{qualification\_criteria}
  \item \texttt{negative\_example\{what\_not\_to\_do, why\_it\_fails\}}
\end{itemize}
\end{tcolorbox}
\caption{Template for failure-driven reusable skill generation.}
\label{prompt:generator}
\end{figure*}